\documentclass[10pt,twocolumn,letterpaper]{article}

\usepackage{cvpr}

\definecolor{cvprblue}{rgb}{0.21,0.49,0.74}
\usepackage[pagebackref,breaklinks,colorlinks,allcolors=cvprblue]{hyperref}

\begin{document}

\title{CoProSketch: Controllable and Progressive Sketch Generation with Diffusion Model}

\author{
    Ruohao Zhan\textsuperscript{1}$^\ast$
    \and
    Yijin Li\textsuperscript{2}$^\ast$
    \and
    Yisheng He\textsuperscript{3}
    \and
    Shuo Chen\textsuperscript{1} 
    \and
    Yichen Shen\textsuperscript{1}
    \and
    Xinyu Chen\textsuperscript{1}
    \and
    Zilong Dong\textsuperscript{3}
    \and
    Zhaoyang Huang\textsuperscript{2}$^\dagger$
    \and
    Guofeng Zhang\textsuperscript{1}$^\dagger$
    \and
    \textsuperscript{1}State Key Laboratory of CAD \& CG, Zhejiang University
    \and
    \textsuperscript{2}Avolution AI
    \and
    \textsuperscript{3}Alibaba Group
}

\maketitle

\footnotetext[1]{Equal contribution.}
\footnotetext[2]{Corresponding authors.}

\begin{figure*}
    \centering
    \includegraphics[width=\textwidth]{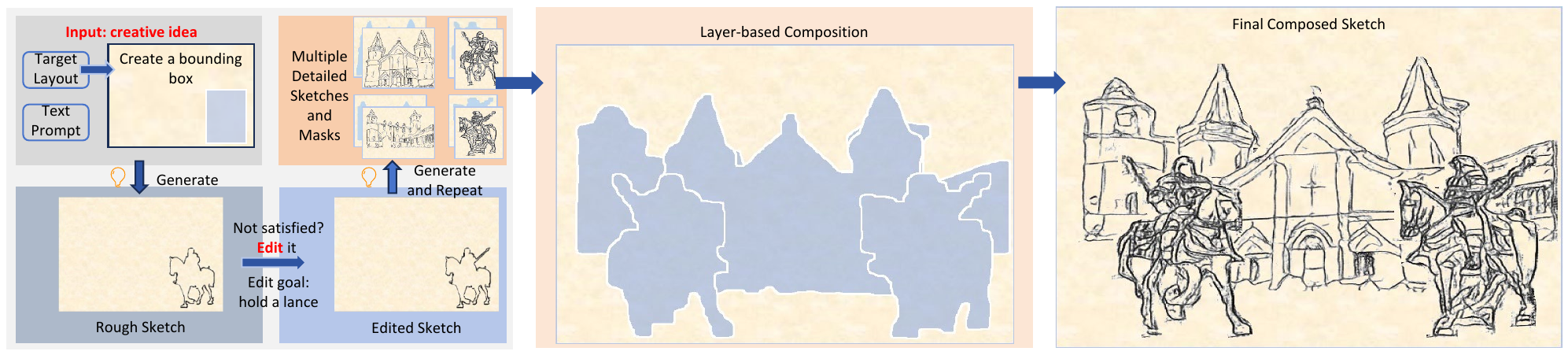}
    \caption{Demonstrations of the proposed pipeline. \textbf{Left}: The proposed pipeline takes a text prompt and an expected layout, represented by a bounding box, as input and generates sketches progressively, from rough to detailed. If the results are unsatisfactory, the user can make timely edits during the rough stage at a low cost. \textbf{Right}: one application is layer-based composition, where the layers (i.e., instance masks) and the sketches are both the output from the proposed pipeline.}
    \label{teaser}
\end{figure*}

\begin{abstract}

Sketches serve as fundamental blueprints in artistic creation because sketch editing is easier and more intuitive than pixel-level RGB image editing for painting artists, yet sketch generation remains unexplored despite advancements in generative models.
We propose a novel framework CoProSketch, providing prominent controllability and details for sketch generation with diffusion models.
A straightforward method is fine-tuning a pretrained image generation diffusion model with binarized sketch images. However, we find that the diffusion models fail to generate clear binary images, which makes the produced sketches chaotic.
We thus propose to represent the sketches by unsigned distance field (UDF), which is continuous and can be easily decoded to sketches through a lightweight network.
With CoProSketch, users generate a rough sketch from a bounding box and a text prompt.
The rough sketch can be manually edited and fed back into the model for iterative refinement and will be decoded to a detailed sketch as the final result.
Additionally, we curate the first large-scale text-sketch paired dataset as the training data. Experiments demonstrate superior semantic consistency and controllability over baselines, offering a practical solution for integrating user feedback into generative workflows.

\end{abstract}

\section{Introduction}

Sketches (or line drawings) are a vital component of the manual painting process, as they capture an artwork’s foundational structure through minimal yet expressive strokes. By focusing on the essential layout, contours and proportions, sketches allow artists to rapidly prototype ideas and refine compositions before committing to fully rendered pieces. Despite this importance, relatively few studies~\cite{xing_diffsketcher_2024,li_photo_sketching_2019,vinker2024sketchagentlanguagedrivensequentialsketch,vinker_clipasso_2022} have addressed the automatic generation of sketches, leaving a notable gap between advancements in research and the practical needs of downstream applications.

On the other hand, recent methods in generative modeling~\cite{yang2023diffusion,review_gan_2023}, particularly diffusion-based approaches~\cite{podell2023sdxlimprovinglatentdiffusion,rombach_high_resolution_2022_CVPR,DDPM_2020}, have made significant strides in generating photorealistic RGB images. However, end-to-end RGB generation remains a challenging task: there is often a gap between synthesized outputs and the complex demands of downstream users. While many conditional generation frameworks~\cite{zhang_adding_2023,ye2023ip} (e.g., semantic maps, reference images) and image post-editing systems~\cite{cao2023masactrl,hertz2022prompt,shi2024dragdiffusion} exist, these systems often rely on a myriad of specialized controls, and adjusting them can become cumbersome for non-expert users. Moreover, these controllable methods still struggle to align crucial requirements such as layout, contours, and proportions. At the same time, manually editing an image in the RGB domain tends to be unintuitive, as it often requires fine-grained pixel-level adjustments, which can be tedious and time-consuming.

Based on these observations, this paper explores a method for generating sketches that better align with user expectations. Our design principles contain: 1) controllability, ensuring that the generated sketches meet user requirements for positioning, size, and shape; and 2) progressive generation, allowing users to pause and adjust the results midway if they are unsatisfied with the intermediate results. The progressive design leverages the inherent editability of sketches.
In particular, as shown in Figure~\ref{teaser}, the proposed pipeline first takes a bounding box and a text prompt as input to generate a rough sketch. If users are not satisfied, they can easily edit the rough result manually. The edited result is then fed forward into the same model (distinguished through a stage indicator) to produce the refined output. Additionally, the proposed pipeline generates an intermediate result, an instance mask of the rough result, after the user confirms the rough stage. This instance mask can be used in layer-based compositing to generate a more complex result on a single canvas.

The most important part of this pipeline is a generative model that outputs a refined sketch given a coarse sketch and a text description.
Unfortunately, previous works~\cite{li_photo_sketching_2019,chan_learning_2022,xing_diffsketcher_2024} on sketch generation provide limited guidance for designing this generative model. Most of them are image-conditioned and primarily focus on transferring the style of an input RGB image into a line drawing. Furthermore, It also means there is a lack of a large-scale text-sketch paired dataset, which is essential for training a robust neural network.

As a result, we refer to recent works~\cite{ani31_2024,ke2024repurposing,long2024wonder3d} that leverage pre-trained diffusion models (e.g., Stable Diffusion XL), which learn strong priors from internet-scale datasets. This approach has the potential to achieve a high level of aesthetic quality and strong semantic consistency, even only fine-tuning the models on our relatively small self-collected dataset. A straightforward idea is to use a pre-trained SDXL~\cite{podell2023sdxlimprovinglatentdiffusion} model and fine-tune directly on the sketch data. However, we empirically find that the results are suboptimal, possibly due to the significant differences in distribution and patterns between RGB images and sketches. Inspired by research in 3D vision~\cite{wang2021neus,liu2023multi}, we introduce the unsigned distance field (UDF) as an intermediate representation for sketches, which provides a more continuous and smooth representation. However, this also introduces a new challenge: decoding the UDF into a sketch. Sketches typically have more complex local ``geometry'' than 3D objects, and as a result, common methods like thresholding or marching squares/cubes~\cite{lorensen1998marching} fail to produce satisfactory sketch results from the UDF. To address this issue, we introduce a lightweight neural network that learns to map from the UDF to the sketch.

To fine-tune the network, we also collect a dataset containing 100k samples. Each sample includes a text description and multi-level contours. The contours start from the coarsest bounding boxes and include object masks, object contours, and detailed textured sketches. 

In summary, our contributions are as follows:
\begin{itemize}
\item We propose a novel sketch generation pipeline that begins with a bounding box and progressively refines the sketch results.
\item We propose a diffusion-based controllable sketch generative model that uses UDF as its representation.
\item We propose a text-sketch paired dataset which contains 100k samples.
\item Extensive experiments demonstrate that the proposed method significantly outperforms the baseline in terms of semantic consistency and controllability. 
\end{itemize}

\begin{figure*}[h]
  \centering
  \includegraphics[width=\linewidth]{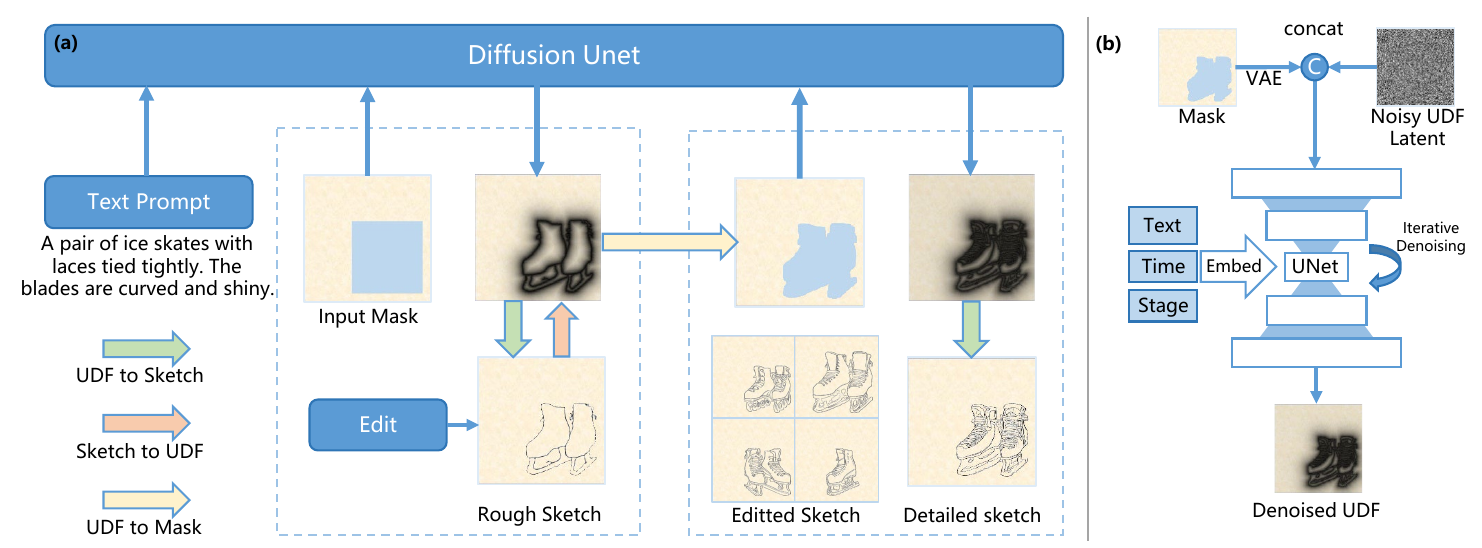}
  \caption{
  \textbf{(a)}: The proposed pipeline begins by taking a text prompt and a rough mask (derived from a bounding box) as input to generate a rough UDF representation. If users find the results unsatisfactory, they have the option to edit the rough result. The rough result, which is the sketch decoded from the UDF, is re-encoded back into the UDF after editing. The edited result is then converted into a instance mask, which is fed back into the same model, guided by a different stage indicator, to produce the refined output.
  \textbf{(b)}: Details of our modified U-Net: The conditional mask is concatenated with the noisy latent. The stage indicator is first converted into an embedding and then added to the time embedding. All other components remain unchanged.
  }
  \label{pipeline}
\end{figure*}

\section{Related work}

\noindent\textbf{Sketch Generation.}
Currently, most sketch generation approaches focus on converting RGB images into sketches, while relatively few address the direct generation of sketches from text prompts. Early edge detection algorithms, such as XDoG\cite{winnemoller2012xdog} and Canny\cite{canny_computational_1986}, extracted object edges from images by calculating pixel gradient information. However, these edge images often include excessive redundant details and lack artistic style, deviating from the core principle of sketches: maintaining simplicity while preserving the semantic essence of objects.

Photo-Sketching\cite{li_photo_sketching_2019} is a deep learning-based approach to sketch generation. It employs a neural network to transform RGB images into sketches, preserving the semantic content of the original images while introducing artistic style and removing redundant details. To train this model, Photo-Sketching constructed a dataset of 5,000 paired RGB images and their corresponding sketches. Similarly, many other methods\cite{Li_2019_Im2Pencil}\cite{Yi_2019_APDrawingGAN} rely on paired RGB-sketch datasets for training. However, the high cost of creating such datasets limits their scale, rendering them inadequate for training modern large-scale models. InformativeDrawings \cite{chan_learning_2022} eliminates the need for paired RGB-sketch data, overcoming this limitation. Instead, it utilizes separate RGB image and sketch datasets, enabling the generation of sketches that match the style of the sketch dataset.

In addition to treating sketch generation as a style transfer task for bitmap data, some methods \cite{vinker_clipasso_2022}\cite{xing_diffsketcher_2024}\cite{Vinker_2023_clipascene}\cite{frans2021clipdrawexploringtexttodrawingsynthesis}\cite{jain2022vectorfusiontexttosvgabstractingpixelbased} represent sketches as a set of Bézier curves with some learnable parameters and use a differentiable renderer to produce sketch images. CLIPasso \cite{vinker_clipasso_2022} requires an RGB image as input to initialize the Bézier curve strokes. By optimizing the semantic loss between the generated sketch and the original RGB image using CLIP \cite{Clip_2021}, it produces high-quality sketches. There are some interesting works that generate handwritten mathematical expressions based on LaTex symbol graphs\cite{Chen_2024_CVPR}.

DiffSketcher \cite{xing_diffsketcher_2024}, a method that can generate sketches directly from text prompts without an RGB image, adopts a similar approach to CLIPasso, replacing the user-provided RGB image with one generated by a diffusion model. Moreover, such methods require re-training for each new input, making the generation of a single sketch time-intensive, often taking several minutes to half an hour. In contrast, methods like InformativeDrawings \cite{chan_learning_2022} can run a neural network inference in less than one second. DALS\cite{kim_dals_2024} is another text-to-sketch method, being able to generate sketches both from text and images. It mainly focuses on landscape sketch, while ours mainly focuses on object sketch. Instead of training a model, SketchAgent\cite{vinker2024sketchagentlanguagedrivensequentialsketch} directly uses a pretrained multimodal LLM to generate sketches based on nature language.

Differing from those image-to-sketch approaches, ours directly generates sketches from text. Compared to existing text-to-sketch approaches, ours is able to generate sketches from rough to detailed, which allows editing at a low cost at rough stage.

\noindent\textbf{Controllable diffusion models.}
Image generation models have achieved remarkable results, and many commercial companies have introduced their own models. However, practitioners in the art-painting industry often require precise control over generated images to fulfill specific needs, rather than allowing the models to generate unconstrained images. ControlNet \cite{zhang_adding_2023} addresses this limitation by enabling control over specific conditions during image generation. For instance, it allows the diffusion model to generate images aligned with object edges by providing an edge map, such as one created using the Canny edge detection algorithm. This functionality supports the creation of images with varying backgrounds, weather conditions, and other attributes while preserving the subject's integrity. Additionally, other approaches \cite{zheng_layoutdiffusion_2024}\cite{cheng_layoutdiffuse_2023}\cite{cheng_hico_2024} offer layout control by accepting several bounding boxes as inputs. These control signals constrain the generated content within predefined areas, allowing users to determine the spatial arrangement of elements on the canvas. ControlNet also supports semantic segmentation as input, enabling users to manage the layout of generated images more effectively.

These approaches focus on directly generating RGB images, which often encounter challenges when dealing with overlapping objects. By taking advantage of the sparse nature of sketches, our approach effectively handles overlapping masks. Instead of generating overlapping content simultaneously, we decouple the overlapping regions and generate them separately, then combine them to produce the final result.

\section{Methodology} \label{introduction}

\subsection{Overview} \label{Overview}

The pipeline of our method is illustrated in Figure \ref{pipeline} (a). We use 2D unsigned distance field(UDF) to represent a sketch. In the first stage, the user needs to provide a bounding box and a text prompt based on their ideas, which serve as inputs to our SDXL-based SketchGenerator module, resulting in a rough UDF within the mask region. In UDF2Sketch module, we use a lightweight network to decode UDFs into sketches. At the same time, UDF2Mask module will extract an instance mask based on the rough UDF. In the second stage, the instance mask and the text prompt serve as inputs to the same SketchGenerator module, resulting in a detailed UDF which can be decoded to detailed sketch by UDF2Sketch module.

To train these modules, we collect a dataset including text description, RGB image, instance mask, bounding box, instance contour and detailed sketch. We will disscuss the details of dataset construction in section \ref{SketchData}.

\subsection{Representation of Sketch: Unsigned Distance Field}  \label{UDF_representation}

In a sketch, the background is typically white, while the foreground is black. Most pixel values in a sketch are 255, while only a few are 0. These abrupt changes in pixel values create sharp contrasts, making it challenging for neural networks to effectively learn from the data. Informed by the widespread application of signed distance fields in 3D vision\cite{Park_2019_DeepSDF}, we adopt a 2D unsigned distance field (UDF) to represent sketches, following \cite{ono_comic_2021}.

\begin{figure}[h]
  \centering
  \includegraphics[width=0.9\linewidth]{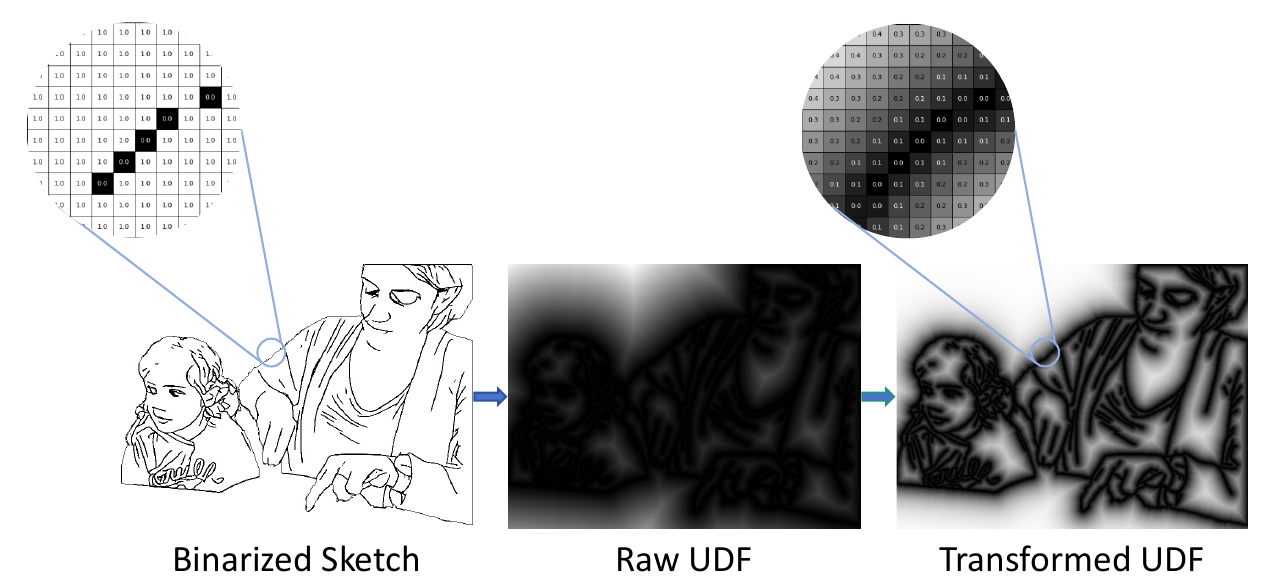}
  \caption{Details of UDF representation. Given a binarized sketch, we compute its UDF representation and transform it by $f(u)$ to adapt it for training networks.}
  \label{dp}
\end{figure}

Figure \ref{dp} illustrates the preprocess steps applied to the raw sketch data before it is fed into the network.
After computing the UDF of the sketches, the information in the sketches becomes blurred.
Therefore, we make a transformation so that the UDF values near the strokes become more distinguishable:
\begin{equation}
  f(u) = 1 - \exp(-u/T),
\end{equation}
where $u$ denotes the input UDF value and $T$ is a hyperparameter. Let's denote the 
width and height of a sketch as $w$ and $h$, following\cite{narita_optical_2019}, we take the value as
\begin{math}
  T = 15\times\sqrt{w^2+h^2}/540
\end{math} on experience.


\subsection{Sketch Generator} \label{SketchGenerator}

Figure \ref{pipeline}(b) shows the network of our SketchGenerator module.

\noindent\textbf{Position Control.}
To ensure that the sketch is generated within the bounding box on the canvas, the mask functions as an effective control signal by providing pixel-level positional information. To incorporate the mask information, we use VAE to encode the mask image into latent, then expand the number of channels in the input convolutional layer of the U-Net from 4 to 8. The original U-Net input latent is concatenated with the mask image latent along the channel dimension, and subsequently fed into the following layers of the network together.

\noindent\textbf{Varying Levels of Abstraction.}
We inject an integer as stage control signal into the network, directing it whether to generate a rough or detailed sketch. We divide the sketch generation process into 3 stages: bounding box, object contour(rough sketch), detailed sketch. When the stage signal is set to 2, it outputs a rough sketch, while a value of 3 triggers the output of a detailed sketch.

Since the stage signal shares similarities with the denoising timestep signal in diffusion denoise process\cite{DDPM_2020}, the structure of our StageEncoder is designed to be identical to that of the TimeEncoder in SDXL. To avoid adding too much new trainable parameters in the U-net, we incorporate time and stage embeddings through summation, which keeps the shape of the embedding tensor unchanged, so that there is no need to modify the layers after this module.

We fully finetune the whole model using our synthetic data with a MSE loss in the framework of DDPM\cite{DDPM_2020}.

\subsection{UDF to Mask} \label{UDF2Mask}

The UDF2Mask module is designed to generate object masks from rough UDFs, which serve as the input to SketchGenerator module to generate detailed sketches. On the other hand, when we want to compose multiple sketches on one canvas, masks can help us handle the occlusion relationship correctly.

It takes the bounding box mask and the rough UDF produced by the SketchGenerator module as inputs. Given the numerous similarities between our target and promptable segmentation, we adopt a network architecture identical to that of SAM\cite{Kirillov_2023_ICCV}. The network comprises three primary modules:
\begin{itemize}
\item ImageEncoder: A ViT-based\cite{alexey2020image} image encoder that encodes the input UDFs into embeddings.
\item MaskEncoder: Encodes the bounding box masks into embeddings.
\item MaskDecoder: Combines the UDF embeddings and box embeddings to generate the final object masks.
\end{itemize}

Since U-Net consumes the majority of computational resources, we adopt a lightweight network architecture, MobileSAM \cite{zhang2023fastersegmentanythinglightweight}, which ensures efficiency while maintaining performance. To train the UDF2Mask module, we incorporated insights from SAM, employing a linear combination of focal loss and dice loss as the final objective:
\begin{equation}
  L = \lambda_f\times L_{focal} + \lambda_d\times L_{dice}.
\end{equation}

\subsection{UDF to Sketch}  \label{UDF2Sketch}

The UDF2Sketch module transforms the generated UDFs into final sketches. There are some traditional methods that can extract isosurface in such a distance field. But in section \ref{Ablation} we illustrate that sketches from these methods lack of aesthetics.

Since UDF-to-sketch task shares a lot of similarity with image-to-sketch, we adopted the design from InformativeDrawing\cite{chan_learning_2022}, which takes an RGB image as input and outputs a high-quality sketch. The network is a simple encoder-decoder architecture with several Res-Net blocks. Following InformativeDrawing, we utilize several loss functions to train it:
\begin{itemize}
\item Adversarial loss: Given a reference image dataset, it encourages the generated sketches to resemble the style of the reference dataset. We choose the Anime Colorization dataset\cite{AnimeSketch} as our target sketch style.
\item Semantic loss: We calculate the CLIP\cite{Clip_2021} embeddings of the raw RGB image and the output sketch, then calculate their MSE loss. Note that our dataset includes (RGB, sketch) pairs.
\item Cycle loss: We follow the settings in InformativeDrawing.
\end{itemize}

\subsection{Sketch Data Engine} \label{SketchData}

\begin{figure}[h]
  \centering
  \includegraphics[width=\linewidth]{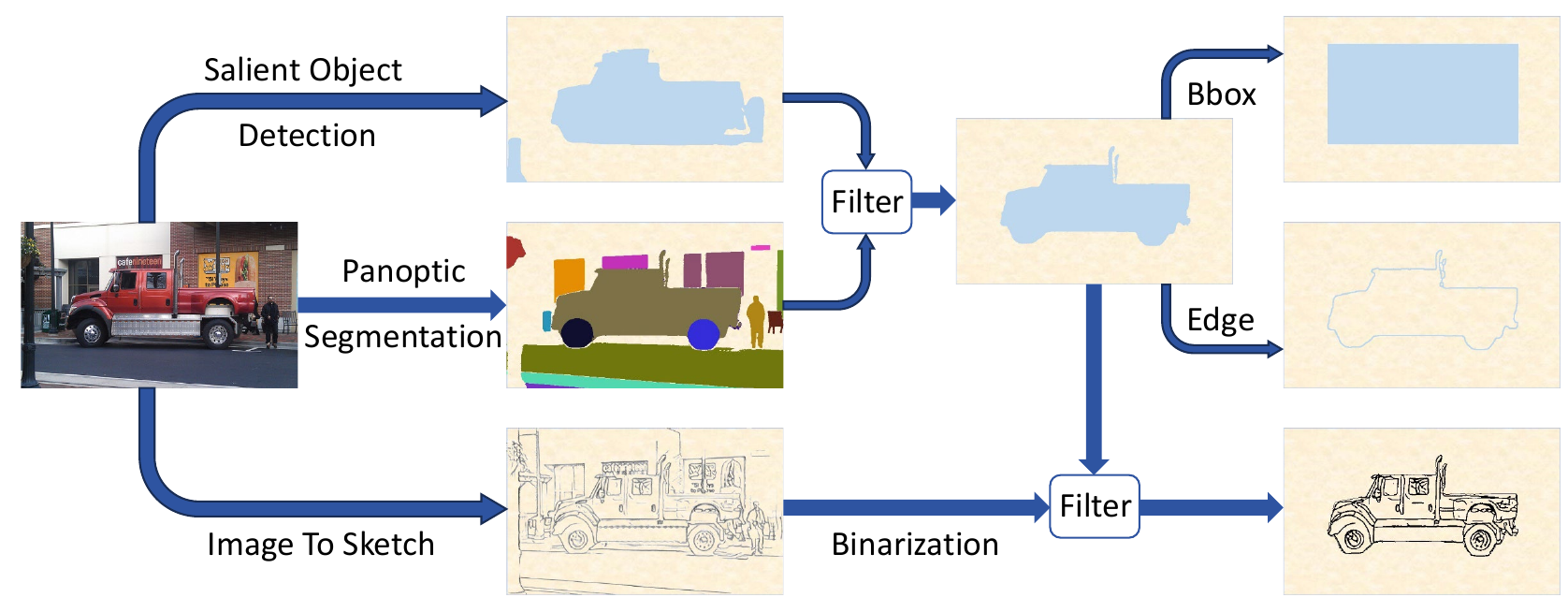}
  \caption{Dataset construction process.}
  \label{ds_pipe}
\end{figure}

Figure \ref{ds_pipe} illustrates the dataset construction pipeline. Starting from a normal RGB image dataset, we construct a new comprehensive sketch dataset. We use Gemini AI to label the RGB images with text caption that describes each image in detail, and then performed a series of image processing steps to obtain the sketch data.

Since scene-level sketches are challenging to be divided into distinct stages based on the degree of abstraction, we focus primarily on object-level sketches.

Given an RGB image, we run a salient object detection (SOD) approach UDASODUPL\cite{yan2022unsupervised} to generate raw salient object masks first. SOD is a computer vision task that identifies and segments the most significant or "salient" objects in an image \cite{wang2024salient}. However, UDASODUPL occasionally fails to fully segment objects, resulting in incomplete or fragmented masks. To address this, we also run SAM2 \cite{ravi2024sam} on the RGB image to obtain more accurate panoptic segmentation results. The regions overlapping between SAM2 and UDASODUPL are identified and combined to produce the final object mask. It's easy to extract the bounding box and object's contour from such mask.

To obtain detailed sketch data, we run an image-to-sketch work InformativeDrawing\cite{chan_learning_2022} on the RGB images, outputing high quality and detailed sketches. To enable the model with positional control, we set the areas outside the object mask to white. To transform the sketch into UDF, we use Otsu's method to binarize the sketch. Although several diffusion-based methods are capable of producing high-quality sketches, we selected InformativeDrawing due to its balance between computational efficiency and sketch quality.

Finally, we apply a filtering process. Images with mask areas that are too small are discarded to ensure the dataset contains only meaningful samples for training.

\subsection{Implementation Details}

To train SketchGenerator module, we utilize the codebase provided by \cite{kohya}. We load the pretrained weights of SDXL, and transfer the style from realistic images into UDF representation on 4 A800 GPUs for 5 epochs.
We use cosine annealing scheduler and the initial learning rate is set to 4.0e-05. The batch size is set to 16 with 3 gradient accumulation steps.
Inference at 512×512 resolution requires about 9 GB of GPU memory, making it feasible to run on personal GPUs.

We build the full dataset based on 3 sub-datasets: COCO2017\cite{lin2014microsoft}, Anime Colorization dataset\cite{AnimeSketch}, SODData. Although COCO2017 contains short text description for the image, longer prompts are more advantageous for the performance of diffusion models\cite{yang2024cogvideoxtexttovideodiffusionmodels}. Therefore, we re-annotated the images using Gemini AI. The Anime Colorization dataset, which includes paired RGB and detailed sketch data, also serves as the training data for InformativeDrawing, allowing us to omit the step of running InformativeDrawing. SODData is a composite dataset formed from several commonly used datasets in SOD tasks, including MSRA\cite{ChengPAMI}, DUT-OMRON\cite{yang2013saliency}, DUTS\cite{wang2017}, HKU-IS\cite{LiYu15}, ECSSD\cite{shi2015hierarchical}, CSSD\cite{yan2013hierarchical}, and SYNSOD\cite{yan2022unsupervised}. These datasets already include salient object masks, allowing us to omit the step of running UDASODUPL. After filtering process, we obtained approximately 100k samples.

\begin{figure*}[h]
  \centering
  \includegraphics[width=\linewidth]{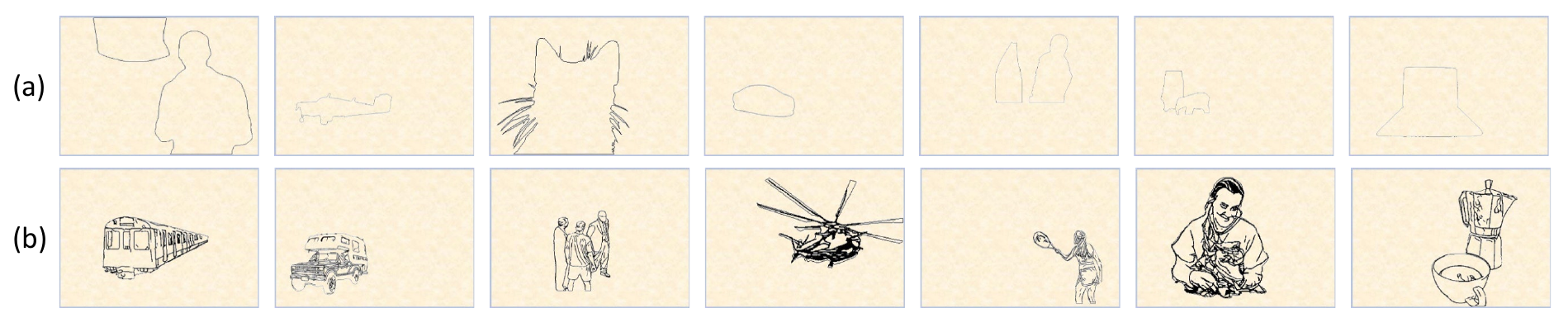}
  \caption{Dataset sample. \textbf{(a)}: Rough sketches. \textbf{(b)}: Detailed sketches.}
  \label{data_sample}
\end{figure*}

Figure \ref{data_sample} shows several samples in the dataset.

\section{Experiments}

\begin{figure*}[h]
  \centering
  \includegraphics[width=0.95\linewidth]{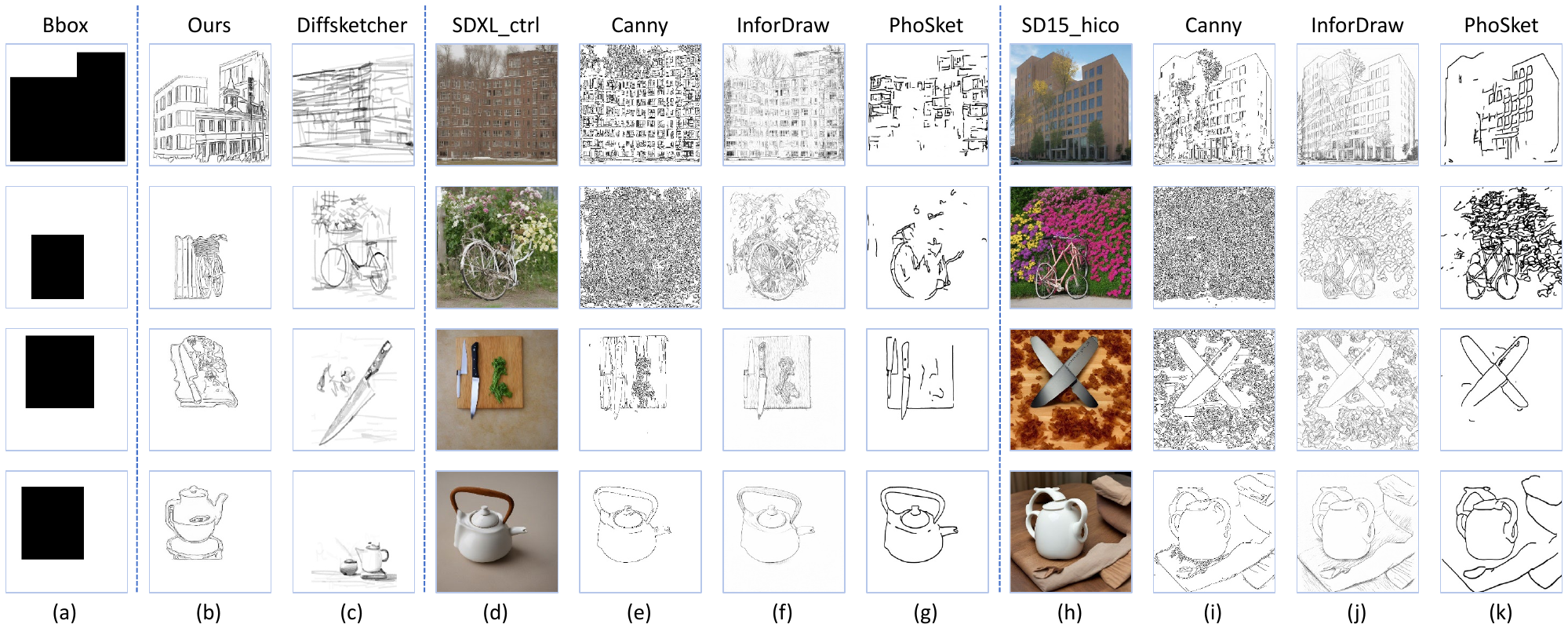}
  \caption{Comparison with baselines. (a - c) represent the box mask input, our results and diffsketcher. (d)(h) represent the RGB images generated by SDXL + ControlNet and SD15+HiCo, and (e - g)(i - k) represent the sketches generated by canny, InformativeDrawings and Photo-Sketching.}
  \label{img_cmp}
\end{figure*}

\subsection{Experimental settings}

DiffSketcher\cite{xing_diffsketcher_2024} is a text-to-sketch framework that shares the same goal as ours, thus we choose it as one of the baselines. Since there are few text-to-sketch works, we also consider a combination of (text-to-image, image-to-sketch) as baselines.

For text-to-image module, we choose 2 approaches: SDXL\cite{podell2023sdxlimprovinglatentdiffusion} with ControlNet\cite{zhang_adding_2023} and SD15\cite{rombach_high_resolution_2022_CVPR} with HiCo\cite{cheng_hico_2024}. ControlNet supports taking semantic segmentation results as input to generate images with expected layout. HiCo focuses on layout control, taking several bounding boxes as input and generating images aligned with it. However, since the SDXL model weights for HiCo have not been released, we can only use SD15 for this comparison.

For image-to-sketch module, we choose 3 approaches: Photo-Sketching\cite{li_photo_sketching_2019}, Canny\cite{canny_computational_1986} and InformativeDrawings\cite{chan_learning_2022}. Among these, InformativeDrawings is used to generate the detailed sketch ground truth for our dataset.

We use ChatGPT to generate about a hundred text prompts as benchmark. Then, we randomly generate a bounding box mask on the canvas for each text prompt to serve as input for position control.

\subsection{Qualitative Evaluation}

Figure \ref{img_cmp} presents a comparison between our method and several baseline approaches. In our method, the generated sketch is successfully confined within the box boundaries, achieving effective position control. DiffSketcher lacks layout control capabilities, which significantly restricts its practicality in real-world applications. For the (text-to-image, image-to-sketch) baseline methods, the intermediate RGB images often contain extensive background details. As a result, the corresponding sketches are cluttered with unnecessary background information, deviating from the fundamental principles of simplicity and abstraction inherent to sketch representation.

The sketches generated by Canny and Photo-Sketching are less visually appealing compared to other methods, often exhibiting broken or inconsistent lines. InformativeDrawings produces superior visual quality, surpassing our approach in terms of aesthetics. However, ControlNet and HiCo exhibit limited layout control, often failing to confine the generated sketch within the input bounding boxes. Our proposed method strikes a balance between robust layout control and high-quality sketch generation, making it well-suited for practical applications.

\subsection{Quantitative Evaluation} \label{Quantitative}

Assessing the quality of generated sketches presents some challenges, as evaluation metrics tailored for RGB images may not be directly applicable to sketches. Following DiffSketcher, we adopted two quantitative metrics for evaluation: aesthetic score and clip semantic score. Aesthetic score is computed using a neural network\cite{aesthetic_predictor_nodate} to predict the aesthetic appeal of images, serving as a proxy for human preference regarding the generated sketches. A higher aesthetic score indicates a stronger alignment with human visual preferences. CLIP score quantifies the semantic consistency between the generated sketch and the text prompt. It is calculated as the cosine similarity between the CLIP\cite{Clip_2021} embeddings of the sketch and the text, with higher scores reflecting better alignment between the sketch and the text prompt.

We observe that the large white areas in sketches significantly impact the above evaluation metrics, often resulting in lower scores. Our method, which strictly confines sketch content within the control boxes, naturally produces images with substantial white areas. In contrast, baseline methods frequently generate sketches that span the entire canvas, resulting in fewer white pixels. To ensure a fair comparison, we resize the baseline sketches to fit within the control boxes, leaving the areas outside the boxes white. This adjustment ensured that the proportion of white pixels in the baseline sketches aligned with those produced by our method, thereby eliminating the influence of differing white areas on the evaluation metrics.

Table \ref{table_cmp} presents the aesthetic scores and CLIP scores of our method compared to the baselines. "T2I" column means the text-to-image method, and "I2S" column means the image-to-sketch method. Our method achieves the highest scores in both aesthetic quality and semantic alignment. These results prove the effectiveness of our approach in generating visually appealing sketches.

\begin{table}
\caption{Evaluation of aesthetic scores and CLIP score.}
\label{table_cmp}
\begin{tabular}{cccc}
    \toprule
T2I & I2S & AES Score & CLIP Score \\
    \midrule
sdxl\_ctrl & Canny & 3.734 & 0.193 \\
sdxl\_ctrl & InforDraw & \underline{4.127}  & \underline{0.227} \\
sdxl\_ctrl & PhoSket & 3.871 & 0.201 \\
\midrule
sd15\_hico & Canny & 3.742 & 0.188 \\
sd15\_hico & InforDraw & 4.078 & 0.210 \\
sd15\_hico & PhoSket & 3.941 & 0.196 \\
\midrule
\multicolumn{2}{c}{Diffsketcher} & 4.043 & 0.217 \\
\midrule
\multicolumn{2}{c}{ours} & \textbf{4.256} & \textbf{0.245} \\
\bottomrule
\end{tabular}
\end{table}

\begin{figure*}[h]
  \centering
  \includegraphics[width=0.8\linewidth]{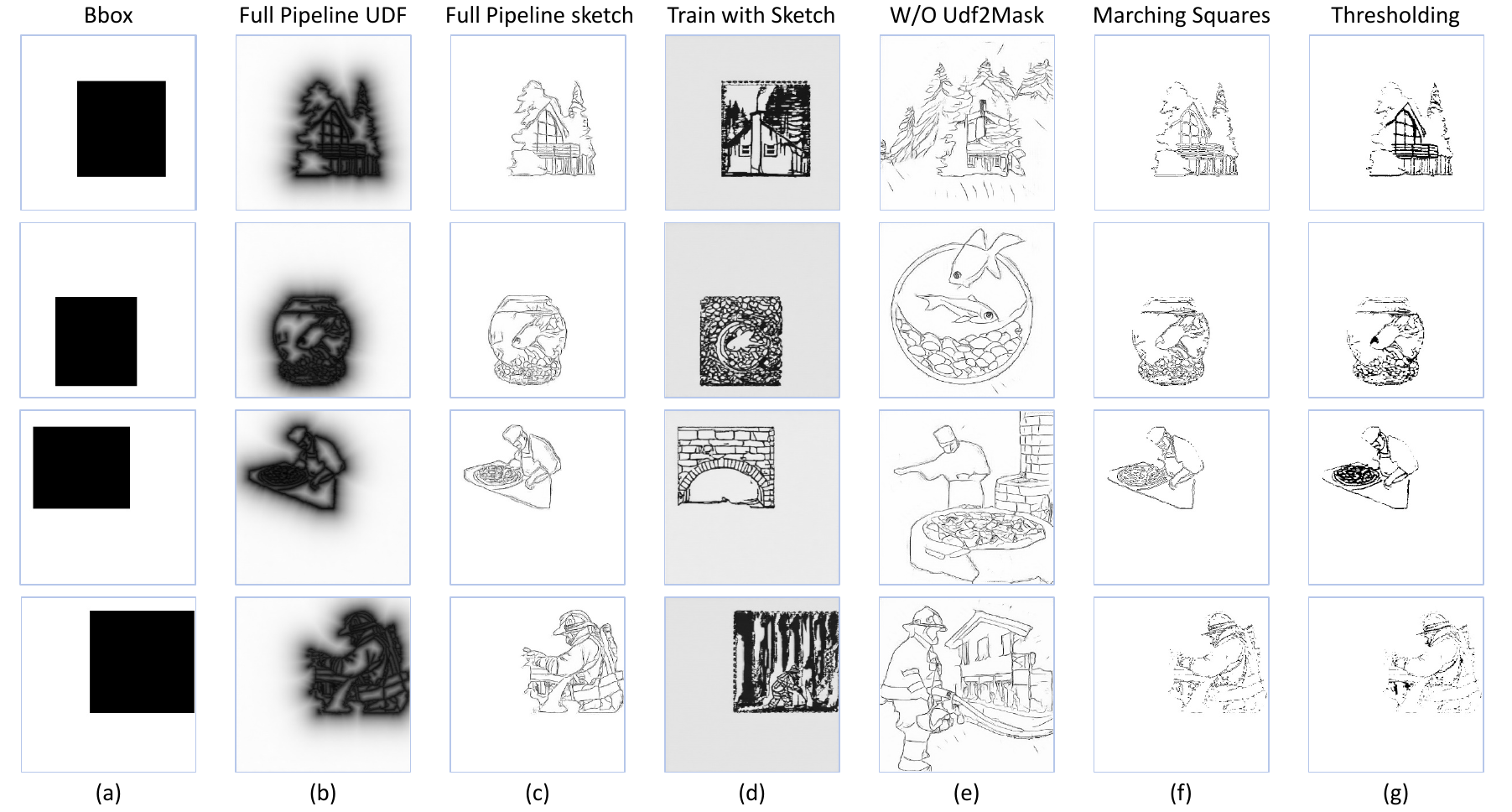}
  \caption{Ablation study. (a - c): bounding box, full pipeline UDF and sketch. (d): Without UDF representation. (e): Without UDF2Mask. (f): Without UDF2Sketch, use marching squares to extract sketch from UDF. (g): Without UDF2Sketch, use thresholding to extract sketch from UDF.}
  \label{ablation_fig}
\end{figure*}

\subsection{User study}

Since the models used to compute aesthetic and CLIP scores are trained on RGB images, a domain gap exists when applying them to sketches. To build a more comprehensive evaluation, we conducted a user study in which volunteers are asked to evaluate the generated sketches based on three dimensions: aesthetic quality, semantic similarity, and positional control accuracy. In order to reduce the workload of volunteers so that they can make more impartial assessments, we only choose (SDXL, ControlNet, InformativeDrawings) and (SD15, Hico, tiveDrawings) as baselines, which get highest scores in aesthetic score and are able to control the layout. We prepared 10 samples from each method and invited 33 volunteers to vote the best between the results of ours and baselines. Aesthetic quality and semantic similarity have the same target as mentioned in section \ref{Quantitative}. Positional control accuracy evaluates how well the sketches adhere to the specified input bounding boxes.

\begin{table}
\caption{User study on three aspects: aesthetic quality, semantic similarity, positional control accuracy. All are measured with voting rates.}
\label{user_study}
\begin{tabular}{ccccc}
    \toprule
T2I & I2S & AES & Semantic & Control \\
    \midrule
sdxl\_ctrl & InforDraw & \textbf{0.385}  & \underline{0.206} & \underline{0.209} \\
\midrule
sd15\_hico & InforDraw & 0.239 & 0.152 & 0.070 \\
\midrule
\multicolumn{2}{c}{ours} & \underline{0.376} & \textbf{0.642} & \textbf{0.721} \\
\bottomrule
\end{tabular}
\end{table}

\begin{table}
\caption{Ablation study.}
\label{ablation}
\begin{tabular}{ccc}
    \toprule
Method & AES Score   & CLIP Score            \\ 
    \midrule
Full pipeline & \textbf{4.256} & \underline{0.245} \\
w/o UDF representation & \underline{4.037} & 0.219 \\
w/o UDF2Mask & 3.849 & 0.218 \\
w/o UDF2Sketch: thresholding & 3.936 & \textbf{0.249} \\
w/o UDF2Sketch: marching square & 3.980 & 0.242 \\
\bottomrule
\end{tabular}
\end{table}

Table \ref{user_study} summarizes the findings from the user study, further validating the superiority of our method. Our approach achieves highest voting rate in position control and semantic similarity. For aesthetic quality, since InformativeDrawings\cite{chan_learning_2022} provide ground truth in our dataset, it's reasonable that voting rate of our results are slightly below the InformativeDrawings's. However, as mentioned in section \ref{Application}, excellent performance in position control and semantic similarity significantly enhanced our practicality.

\subsection{Ablation Study} \label{Ablation}

\begin{figure*}[h]
  \centering
  \includegraphics[width=0.9\linewidth]{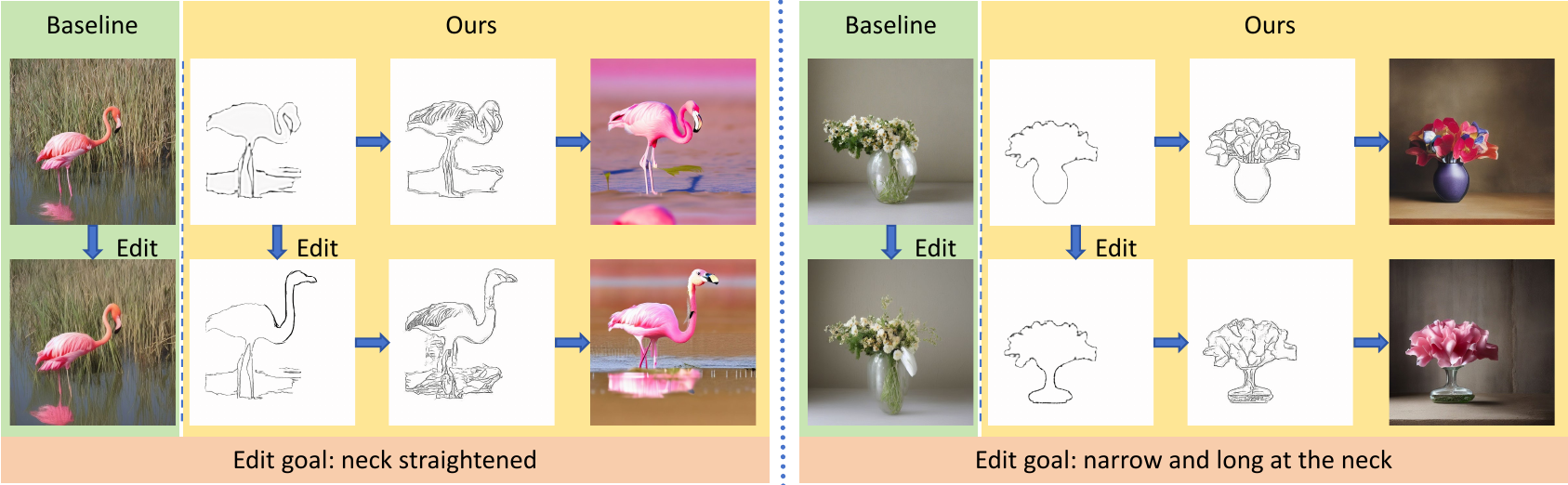}
  \caption{Comparison between editing in text-to-RGB pipeline(baseline) and text-rough-detailed-RGB pipeline(ours). Our pipeline shows better editing performance.}
  \label{edit}
\end{figure*}

\begin{figure*}[h]
  \centering
  \includegraphics[width=0.9\linewidth]{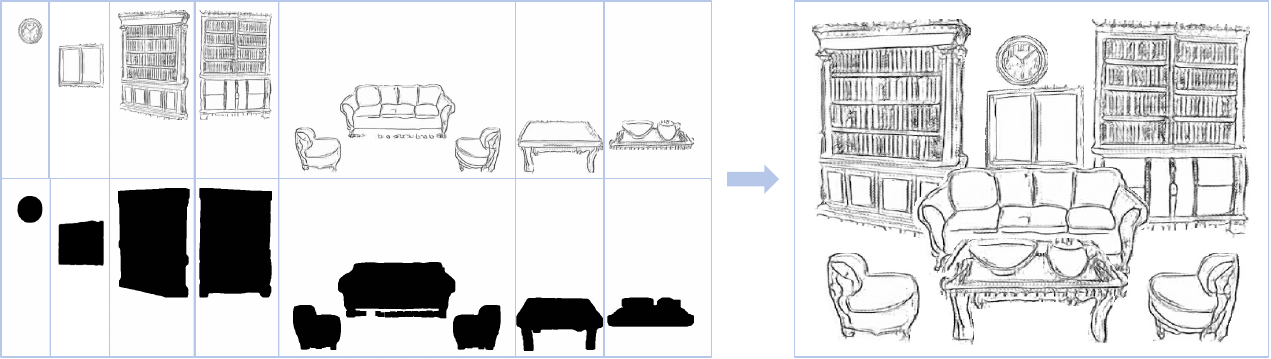}
  \caption{Application of composing multiple objects into one sketch.}
  \label{composite}
\end{figure*}

We conduct an ablation study focusing on three key modules in our pipeline: \textbf{UDF representation}, \textbf{UDF2Mask}, and \textbf{UDF2Sketch}. Figure \ref{ablation_fig} qualitatively illustrates the effect of excluding each module from the pipeline. Table \ref{ablation} presents the quantitative evaluation metrics discussed in Section \ref{Quantitative}.

\noindent\textbf{UDF representation.}
For comparison, in the absence of UDF representation, sketches are used directly as ground truth to train the diffusion model. The results in Figure \ref{ablation_fig} indicate that training directly with sketches often results in large color blocks and a significant decline in image quality. Table \ref{ablation} reveals that the generated sketch maintains little semantic consistency with the text prompt. This highlights the importance of UDF representation to ensure the generation of high-quality sketches.

\noindent\textbf{UDF2Mask.}
The UDF2Mask module is critical for precise positional control. Without this module, we use stage 2 sketch as the control signal for stage 3, which significantly increasing the difficulty of maintaining accurate positional control. Figure \ref{ablation_fig} reveals that there are more instances where sketches exceed the boundaries of the bounding box without UDF2Mask module. It will be hard to achieve bounding box control without this module.

\noindent\textbf{UDF2Sketch.}
The UDF2Sketch module plays an essential role in the quality of the final output sketch. Figure \ref{ablation_fig} provides a comparison between sketches generated using our UDF2Sketch module and those produced by traditional approaches: marching squares and thresholding. Marching squares is widely used for extracting isosurfaces from 2D matrix data. Thresholding directly sets pixel values below a specific threshold to black, while leaving other pixels white. Our network-based method delivers higher-quality sketches with smoother lines, while traditional approaches often result in fragmented sketches that lack visual appeal.

\subsection{Application} \label{Application}

\noindent\textbf{Editability.}
sketches can serve as the input to ControlNet\cite{zhang_adding_2023}, guiding the generation of RGB images. It is challenging to directly edit an RGB image when unsatisfied. However, our approach enables the progressive generation of sketches, allowing for timely edits during the creative process when unsatisfied with the results. To validate this capability, we compare the results of editing generated images in two image generation pipelines. Baseline: We use SDXL\cite{podell2023sdxlimprovinglatentdiffusion} with ControlNet\cite{zhang_adding_2023} to implement a text-to-RGB pipeline, followed by editing the RGB image using MasaCtrl\cite{cao2023masactrl}. Ours: After generating the rough sketch, it is manually edited, and the modified version is used to generate the detailed sketch, which is used to generate RGB image by ControlNet. Figure \ref{edit} illustrates the results of these experiments. Editing RGB images was significantly challenging in baseline, resulting in unsatisfactory editing results. In contrast, our pipeline allows for easier editing at the stage 2 sketch level, owing to the sparse representation of sketches.

\noindent\textbf{Complex layout.}
Generating a sketch with complex layout in one prompt is challenging. It's hard for model to understand the relative position relationship and occlusion relationship, resulting in unsatisfied sketches. In our approach, users can generate each component individually and finally compose them on one canvas based on the mask. Figure \ref{composite} shows that the user can easily overlay foreground objects onto background objects on the canvas, resulting in a complex layout with correct occlusion relationship.

\section{Discussion and Conclusions}

This paper introduces a novel pipeline for controllable and progressively refined sketch generation, addressing the crucial need for user-friendly creative tools. The key to this pipeline is a diffusion-based controllable sketch generative model that uses UDF as its representation. To train the network, this paper also collects the first text-sketch paired dataset. Experimental results indicate that the proposed method exceeds the baselines in semantic alignment and controllability.
One limitation is that the aesthetic performance is limited by the image-to-sketch approach used for data construction.

\clearpage
\clearpage

{
\small
\bibliographystyle{ieeenat_fullname}
\bibliography{references}

\begin{thebibliography}{54}
\providecommand{\natexlab}[1]{#1}
\providecommand{\url}[1]{\texttt{#1}}
\expandafter\ifx\csname urlstyle\endcsname\relax
  \providecommand{\doi}[1]{doi: #1}\else
  \providecommand{\doi}{doi: \begingroup \urlstyle{rm}\Url}\fi

\bibitem[Alexey(2020)]{alexey2020image}
Dosovitskiy Alexey.
\newblock An image is worth 16x16 words: Transformers for image recognition at scale.
\newblock \emph{arXiv preprint arXiv: 2010.11929}, 2020.

\bibitem[cagliostrolab(2024)]{ani31_2024}
cagliostrolab.
\newblock cagliostrolab/animagine-xl-3.1 · {Hugging} {Face}, 2024.

\bibitem[Canny(1986)]{canny_computational_1986}
John Canny.
\newblock A {Computational} {Approach} to {Edge} {Detection}.
\newblock \emph{IEEE Transactions on Pattern Analysis and Machine Intelligence}, PAMI-8\penalty0 (6):\penalty0 679--698, 1986.
\newblock Conference Name: IEEE Transactions on Pattern Analysis and Machine Intelligence.

\bibitem[Cao et~al.(2023)Cao, Wang, Qi, Shan, Qie, and Zheng]{cao2023masactrl}
Mingdeng Cao, Xintao Wang, Zhongang Qi, Ying Shan, Xiaohu Qie, and Yinqiang Zheng.
\newblock Masactrl: Tuning-free mutual self-attention control for consistent image synthesis and editing.
\newblock In \emph{Proceedings of the IEEE/CVF International Conference on Computer Vision}, pages 22560--22570, 2023.

\bibitem[Chan et~al.(2022)Chan, Durand, and Isola]{chan_learning_2022}
Caroline Chan, Frédo Durand, and Phillip Isola.
\newblock Learning {To} {Generate} {Line} {Drawings} {That} {Convey} {Geometry} and {Semantics}.
\newblock pages 7915--7925, 2022.

\bibitem[Chen et~al.(2024)Chen, Gao, Zhang, Qiao, and Wang]{Chen_2024_CVPR}
Yu Chen, Fei Gao, Yanguang Zhang, Maoying Qiao, and Nannan Wang.
\newblock Generating handwritten mathematical expressions from symbol graphs: An end-to-end pipeline.
\newblock In \emph{Proceedings of the IEEE/CVF Conference on Computer Vision and Pattern Recognition (CVPR)}, pages 15675--15685, 2024.

\bibitem[Cheng et~al.(2024)Cheng, Ma, Wu, Liu, Ma, Wu, Leng, and Yin]{cheng_hico_2024}
Bo Cheng, Yuhang Ma, Liebucha Wu, Shanyuan Liu, Ao Ma, Xiaoyu Wu, Dawei Leng, and Yuhui Yin.
\newblock {HiCo}: {Hierarchical} {Controllable} {Diffusion} {Model} for {Layout}-to-image {Generation}, 2024.
\newblock arXiv:2410.14324 [cs].

\bibitem[Cheng et~al.(2023)Cheng, Liang, Shi, He, Xiao, and Li]{cheng_layoutdiffuse_2023}
Jiaxin Cheng, Xiao Liang, Xingjian Shi, Tong He, Tianjun Xiao, and Mu Li.
\newblock {LayoutDiffuse}: {Adapting} {Foundational} {Diffusion} {Models} for {Layout}-to-{Image} {Generation}, 2023.
\newblock arXiv:2302.08908 [cs].

\bibitem[Cheng et~al.(2015)Cheng, Mitra, Huang, Torr, and Hu]{ChengPAMI}
Ming-Ming Cheng, Niloy~J. Mitra, Xiaolei Huang, Philip H.~S. Torr, and Shi-Min Hu.
\newblock Global contrast based salient region detection.
\newblock \emph{IEEE TPAMI}, 37\penalty0 (3):\penalty0 569--582, 2015.

\bibitem[Frans et~al.(2021)Frans, Soros, and Witkowski]{frans2021clipdrawexploringtexttodrawingsynthesis}
Kevin Frans, L.~B. Soros, and Olaf Witkowski.
\newblock Clipdraw: Exploring text-to-drawing synthesis through language-image encoders, 2021.

\bibitem[Hertz et~al.(2022)Hertz, Mokady, Tenenbaum, Aberman, Pritch, and Cohen-Or]{hertz2022prompt}
Amir Hertz, Ron Mokady, Jay Tenenbaum, Kfir Aberman, Yael Pritch, and Daniel Cohen-Or.
\newblock Prompt-to-prompt image editing with cross attention control.
\newblock \emph{arXiv preprint arXiv:2208.01626}, 2022.

\bibitem[Ho et~al.(2020)Ho, Jain, and Abbeel]{DDPM_2020}
Jonathan Ho, Ajay Jain, and Pieter Abbeel.
\newblock Denoising diffusion probabilistic models.
\newblock In \emph{Advances in Neural Information Processing Systems}, pages 6840--6851. Curran Associates, Inc., 2020.

\bibitem[Jain et~al.(2022)Jain, Xie, and Abbeel]{jain2022vectorfusiontexttosvgabstractingpixelbased}
Ajay Jain, Amber Xie, and Pieter Abbeel.
\newblock Vectorfusion: Text-to-svg by abstracting pixel-based diffusion models, 2022.

\bibitem[Ke et~al.(2024)Ke, Obukhov, Huang, Metzger, Daudt, and Schindler]{ke2024repurposing}
Bingxin Ke, Anton Obukhov, Shengyu Huang, Nando Metzger, Rodrigo~Caye Daudt, and Konrad Schindler.
\newblock Repurposing diffusion-based image generators for monocular depth estimation.
\newblock In \emph{Proceedings of the IEEE/CVF Conference on Computer Vision and Pattern Recognition}, pages 9492--9502, 2024.

\bibitem[Kim et~al.(2024)Kim, Yang, and Min]{kim_dals_2024}
Junho Kim, Heekyung Yang, and Kyungha Min.
\newblock {DALS}: {Diffusion}-{Based} {Artistic} {Landscape} {Sketch}.
\newblock \emph{Mathematics}, 12\penalty0 (2):\penalty0 238, 2024.

\bibitem[Kim(2018)]{AnimeSketch}
Taebum Kim.
\newblock Anime {Sketch} {Colorization} {Pair}, 2018.

\bibitem[Kirillov et~al.(2023)Kirillov, Mintun, Ravi, Mao, Rolland, Gustafson, Xiao, Whitehead, Berg, Lo, Dollar, and Girshick]{Kirillov_2023_ICCV}
Alexander Kirillov, Eric Mintun, Nikhila Ravi, Hanzi Mao, Chloe Rolland, Laura Gustafson, Tete Xiao, Spencer Whitehead, Alexander~C. Berg, Wan-Yen Lo, Piotr Dollar, and Ross Girshick.
\newblock Segment anything.
\newblock In \emph{Proceedings of the IEEE/CVF International Conference on Computer Vision (ICCV)}, pages 4015--4026, 2023.

\bibitem[Kohya(2024)]{kohya}
Kohya.
\newblock kohya-ss/sd-scripts, 2024.

\bibitem[Li and Yu(2015)]{LiYu15}
G. Li and Y. Yu.
\newblock Visual saliency based on multiscale deep features.
\newblock In \emph{IEEE Conference on Computer Vision and Pattern Recognition (CVPR)}, pages 5455--5463, 2015.

\bibitem[Li et~al.(2019{\natexlab{a}})Li, Lin, Mech, Yumer, and Ramanan]{li_photo_sketching_2019}
Mengtian Li, Zhe Lin, Radomir Mech, Ersin Yumer, and Deva Ramanan.
\newblock Photo-sketching: Inferring contour drawings from images.
\newblock In \emph{2019 IEEE Winter Conference on Applications of Computer Vision (WACV)}, pages 1403--1412, 2019{\natexlab{a}}.

\bibitem[Li et~al.(2019{\natexlab{b}})Li, Fang, Hertzmann, Shechtman, and Yang]{Li_2019_Im2Pencil}
Yijun Li, Chen Fang, Aaron Hertzmann, Eli Shechtman, and Ming-Hsuan Yang.
\newblock Im2pencil: Controllable pencil illustration from photographs.
\newblock In \emph{Proceedings of the IEEE/CVF Conference on Computer Vision and Pattern Recognition (CVPR)}, 2019{\natexlab{b}}.

\bibitem[Lin et~al.(2014)Lin, Maire, Belongie, Hays, Perona, Ramanan, Doll{\'a}r, and Zitnick]{lin2014microsoft}
Tsung-Yi Lin, Michael Maire, Serge Belongie, James Hays, Pietro Perona, Deva Ramanan, Piotr Doll{\'a}r, and C~Lawrence Zitnick.
\newblock Microsoft coco: Common objects in context.
\newblock In \emph{European conference on computer vision}, pages 740--755. Springer, 2014.

\bibitem[Liu et~al.(2023)Liu, Li, Teng, Bao, Zhang, Zhang, and Cui]{liu2023multi}
Xinyang Liu, Yijin Li, Yanbin Teng, Hujun Bao, Guofeng Zhang, Yinda Zhang, and Zhaopeng Cui.
\newblock Multi-modal neural radiance field for monocular dense slam with a light-weight tof sensor.
\newblock In \emph{Proceedings of the ieee/cvf international conference on computer vision}, pages 1--11, 2023.

\bibitem[Long et~al.(2024)Long, Guo, Lin, Liu, Dou, Liu, Ma, Zhang, Habermann, Theobalt, et~al.]{long2024wonder3d}
Xiaoxiao Long, Yuan-Chen Guo, Cheng Lin, Yuan Liu, Zhiyang Dou, Lingjie Liu, Yuexin Ma, Song-Hai Zhang, Marc Habermann, Christian Theobalt, et~al.
\newblock Wonder3d: Single image to 3d using cross-domain diffusion.
\newblock In \emph{Proceedings of the IEEE/CVF Conference on Computer Vision and Pattern Recognition}, pages 9970--9980, 2024.

\bibitem[Lorensen and Cline(1998)]{lorensen1998marching}
William~E Lorensen and Harvey~E Cline.
\newblock Marching cubes: A high resolution 3d surface construction algorithm.
\newblock In \emph{Seminal graphics: pioneering efforts that shaped the field}, pages 347--353. 1998.

\bibitem[Narita et~al.(2019)Narita, Hirakawa, and Aizawa]{narita_optical_2019}
Rei Narita, Keigo Hirakawa, and Kiyoharu Aizawa.
\newblock Optical {Flow} {Based} {Line} {Drawing} {Frame} {Interpolation} {Using} {Distance} {Transform} to {Support} {Inbetweenings}.
\newblock In \emph{2019 {IEEE} {International} {Conference} on {Image} {Processing} ({ICIP})}, pages 4200--4204, Taipei, Taiwan, 2019. IEEE.

\bibitem[Ono et~al.(2021)Ono, Aizawa, and Matsui]{ono_comic_2021}
Naoki Ono, Kiyoharu Aizawa, and Yusuke Matsui.
\newblock Comic {Image} {Inpainting} via {Distance} {Transform}.
\newblock In \emph{{SIGGRAPH} {Asia} 2021 {Technical} {Communications}}, pages 1--4, Tokyo Japan, 2021. ACM.

\bibitem[Park et~al.(2019)Park, Florence, Straub, Newcombe, and Lovegrove]{Park_2019_DeepSDF}
Jeong~Joon Park, Peter Florence, Julian Straub, Richard Newcombe, and Steven Lovegrove.
\newblock Deepsdf: Learning continuous signed distance functions for shape representation.
\newblock In \emph{Proceedings of the IEEE/CVF Conference on Computer Vision and Pattern Recognition (CVPR)}, 2019.

\bibitem[Podell et~al.(2023)Podell, English, Lacey, Blattmann, Dockhorn, Müller, Penna, and Rombach]{podell2023sdxlimprovinglatentdiffusion}
Dustin Podell, Zion English, Kyle Lacey, Andreas Blattmann, Tim Dockhorn, Jonas Müller, Joe Penna, and Robin Rombach.
\newblock Sdxl: Improving latent diffusion models for high-resolution image synthesis, 2023.

\bibitem[Radford et~al.(2021)Radford, Kim, Hallacy, Ramesh, Goh, Agarwal, Sastry, Askell, Mishkin, Clark, Krueger, and Sutskever]{Clip_2021}
Alec Radford, Jong~Wook Kim, Chris Hallacy, Aditya Ramesh, Gabriel Goh, Sandhini Agarwal, Girish Sastry, Amanda Askell, Pamela Mishkin, Jack Clark, Gretchen Krueger, and Ilya Sutskever.
\newblock Learning transferable visual models from natural language supervision.
\newblock In \emph{Proceedings of the 38th International Conference on Machine Learning}, pages 8748--8763. PMLR, 2021.

\bibitem[Ravi et~al.(2024)Ravi, Gabeur, Hu, Hu, Ryali, Ma, Khedr, R{\"a}dle, Rolland, Gustafson, et~al.]{ravi2024sam}
Nikhila Ravi, Valentin Gabeur, Yuan-Ting Hu, Ronghang Hu, Chaitanya Ryali, Tengyu Ma, Haitham Khedr, Roman R{\"a}dle, Chloe Rolland, Laura Gustafson, et~al.
\newblock Sam 2: Segment anything in images and videos.
\newblock \emph{arXiv preprint arXiv:2408.00714}, 2024.

\bibitem[Rombach et~al.(2022)Rombach, Blattmann, Lorenz, Esser, and Ommer]{rombach_high_resolution_2022_CVPR}
Robin Rombach, Andreas Blattmann, Dominik Lorenz, Patrick Esser, and Bj\"orn Ommer.
\newblock High-resolution image synthesis with latent diffusion models.
\newblock In \emph{Proceedings of the IEEE/CVF Conference on Computer Vision and Pattern Recognition (CVPR)}, pages 10684--10695, 2022.

\bibitem[Schuhmann(2022)]{aesthetic_predictor_nodate}
Christoph Schuhmann.
\newblock christophschuhmann/improved-aesthetic-predictor: {CLIP}+{MLP} {Aesthetic} {Score} {Predictor}, 2022.

\bibitem[Shi et~al.(2015)Shi, Yan, Xu, and Jia]{shi2015hierarchical}
Jianping Shi, Qiong Yan, Li Xu, and Jiaya Jia.
\newblock Hierarchical image saliency detection on extended cssd.
\newblock \emph{IEEE transactions on pattern analysis and machine intelligence}, 38\penalty0 (4):\penalty0 717--729, 2015.

\bibitem[Shi et~al.(2024)Shi, Xue, Liew, Pan, Yan, Zhang, Tan, and Bai]{shi2024dragdiffusion}
Yujun Shi, Chuhui Xue, Jun~Hao Liew, Jiachun Pan, Hanshu Yan, Wenqing Zhang, Vincent~YF Tan, and Song Bai.
\newblock Dragdiffusion: Harnessing diffusion models for interactive point-based image editing.
\newblock In \emph{Proceedings of the IEEE/CVF Conference on Computer Vision and Pattern Recognition}, pages 8839--8849, 2024.

\bibitem[{Trevisan de Souza} et~al.(2023){Trevisan de Souza}, Marques, Batagelo, and Gois]{review_gan_2023}
Vinicius~Luis {Trevisan de Souza}, Bruno Augusto~Dorta Marques, Harlen~Costa Batagelo, and João~Paulo Gois.
\newblock A review on generative adversarial networks for image generation.
\newblock \emph{Computers \& Graphics}, 114:\penalty0 13--25, 2023.

\bibitem[Vinker et~al.(2022)Vinker, Pajouheshgar, Bo, Bachmann, Bermano, Cohen-Or, Zamir, and Shamir]{vinker_clipasso_2022}
Yael Vinker, Ehsan Pajouheshgar, Jessica~Y. Bo, Roman~Christian Bachmann, Amit~Haim Bermano, Daniel Cohen-Or, Amir Zamir, and Ariel Shamir.
\newblock {CLIPasso}: {Semantically}-{Aware} {Object} {Sketching}, 2022.
\newblock arXiv:2202.05822.

\bibitem[Vinker et~al.(2023)Vinker, Alaluf, Cohen-Or, and Shamir]{Vinker_2023_clipascene}
Yael Vinker, Yuval Alaluf, Daniel Cohen-Or, and Ariel Shamir.
\newblock Clipascene: Scene sketching with different types and levels of abstraction.
\newblock In \emph{Proceedings of the IEEE/CVF International Conference on Computer Vision (ICCV)}, pages 4146--4156, 2023.

\bibitem[Vinker et~al.(2024)Vinker, Shaham, Zheng, Zhao, Fan, and Torralba]{vinker2024sketchagentlanguagedrivensequentialsketch}
Yael Vinker, Tamar~Rott Shaham, Kristine Zheng, Alex Zhao, Judith~E Fan, and Antonio Torralba.
\newblock Sketchagent: Language-driven sequential sketch generation, 2024.

\bibitem[Wang et~al.(2017)Wang, Lu, Wang, Feng, Wang, Yin, and Ruan]{wang2017}
Lijun Wang, Huchuan Lu, Yifan Wang, Mengyang Feng, Dong Wang, Baocai Yin, and Xiang Ruan.
\newblock Learning to detect salient objects with image-level supervision.
\newblock In \emph{CVPR}, 2017.

\bibitem[Wang et~al.(2021)Wang, Liu, Liu, Theobalt, Komura, and Wang]{wang2021neus}
Peng Wang, Lingjie Liu, Yuan Liu, Christian Theobalt, Taku Komura, and Wenping Wang.
\newblock Neus: Learning neural implicit surfaces by volume rendering for multi-view reconstruction.
\newblock \emph{arXiv preprint arXiv:2106.10689}, 2021.

\bibitem[Wang et~al.(2024)Wang, Yu, Lim, and Wong]{wang2024salient}
Xiuwenxin Wang, Siyue Yu, Eng~Gee Lim, and ML~Dennis Wong.
\newblock Salient object detection: a mini review.
\newblock \emph{Frontiers in Signal Processing}, 4:\penalty0 1356793, 2024.

\bibitem[Winnem{\"o}ller et~al.(2012)Winnem{\"o}ller, Kyprianidis, and Olsen]{winnemoller2012xdog}
Holger Winnem{\"o}ller, Jan~Eric Kyprianidis, and Sven~C Olsen.
\newblock Xdog: An extended difference-of-gaussians compendium including advanced image stylization.
\newblock \emph{Computers \& Graphics}, 36\penalty0 (6):\penalty0 740--753, 2012.

\bibitem[Xing et~al.(2024)Xing, Wang, Zhou, Zhang, Yu, and Xu]{xing_diffsketcher_2024}
Ximing Xing, Chuang Wang, Haitao Zhou, Jing Zhang, Qian Yu, and Dong Xu.
\newblock {DiffSketcher}: {Text} {Guided} {Vector} {Sketch} {Synthesis} through {Latent} {Diffusion} {Models}, 2024.
\newblock arXiv:2306.14685 [cs].

\bibitem[Yan et~al.(2022)Yan, Wu, Liu, Zeng, Lin, and Li]{yan2022unsupervised}
Pengxiang Yan, Ziyi Wu, Mengmeng Liu, Kun Zeng, Liang Lin, and Guanbin Li.
\newblock Unsupervised domain adaptive salient object detection through uncertainty-aware pseudo-label learning.
\newblock In \emph{Proceedings of the AAAI Conference on Artificial Intelligence}, pages 3000--3008, 2022.

\bibitem[Yan et~al.(2013)Yan, Xu, Shi, and Jia]{yan2013hierarchical}
Qiong Yan, Li Xu, Jianping Shi, and Jiaya Jia.
\newblock Hierarchical saliency detection.
\newblock In \emph{Proceedings of the IEEE conference on computer vision and pattern recognition}, pages 1155--1162, 2013.

\bibitem[Yang et~al.(2013)Yang, Zhang, Lu, Ruan, and Yang]{yang2013saliency}
Chuan Yang, Lihe Zhang, Huchuan Lu, Xiang Ruan, and Ming-Hsuan Yang.
\newblock Saliency detection via graph-based manifold ranking.
\newblock In \emph{2013 IEEE Conference on Computer Vision and Pattern Recognition}, pages 3166--3173, 2013.

\bibitem[Yang et~al.(2023)Yang, Zhang, Song, Hong, Xu, Zhao, Zhang, Cui, and Yang]{yang2023diffusion}
Ling Yang, Zhilong Zhang, Yang Song, Shenda Hong, Runsheng Xu, Yue Zhao, Wentao Zhang, Bin Cui, and Ming-Hsuan Yang.
\newblock Diffusion models: A comprehensive survey of methods and applications.
\newblock \emph{ACM Computing Surveys}, 56\penalty0 (4):\penalty0 1--39, 2023.

\bibitem[Yang et~al.(2024)Yang, Teng, Zheng, Ding, Huang, Xu, Yang, Hong, Zhang, Feng, Yin, Gu, Zhang, Wang, Cheng, Liu, Xu, Dong, and Tang]{yang2024cogvideoxtexttovideodiffusionmodels}
Zhuoyi Yang, Jiayan Teng, Wendi Zheng, Ming Ding, Shiyu Huang, Jiazheng Xu, Yuanming Yang, Wenyi Hong, Xiaohan Zhang, Guanyu Feng, Da Yin, Xiaotao Gu, Yuxuan Zhang, Weihan Wang, Yean Cheng, Ting Liu, Bin Xu, Yuxiao Dong, and Jie Tang.
\newblock Cogvideox: Text-to-video diffusion models with an expert transformer, 2024.

\bibitem[Ye et~al.(2023)Ye, Zhang, Liu, Han, and Yang]{ye2023ip}
Hu Ye, Jun Zhang, Sibo Liu, Xiao Han, and Wei Yang.
\newblock Ip-adapter: Text compatible image prompt adapter for text-to-image diffusion models.
\newblock \emph{arXiv preprint arXiv:2308.06721}, 2023.

\bibitem[Yi et~al.(2019)Yi, Liu, Lai, and Rosin]{Yi_2019_APDrawingGAN}
Ran Yi, Yong-Jin Liu, Yu-Kun Lai, and Paul~L. Rosin.
\newblock Apdrawinggan: Generating artistic portrait drawings from face photos with hierarchical gans.
\newblock In \emph{Proceedings of the IEEE/CVF Conference on Computer Vision and Pattern Recognition (CVPR)}, 2019.

\bibitem[Zhang et~al.(2023{\natexlab{a}})Zhang, Han, Qiao, Kim, Bae, Lee, and Hong]{zhang2023fastersegmentanythinglightweight}
Chaoning Zhang, Dongshen Han, Yu Qiao, Jung~Uk Kim, Sung-Ho Bae, Seungkyu Lee, and Choong~Seon Hong.
\newblock Faster segment anything: Towards lightweight sam for mobile applications, 2023{\natexlab{a}}.

\bibitem[Zhang et~al.(2023{\natexlab{b}})Zhang, Rao, and Agrawala]{zhang_adding_2023}
Lvmin Zhang, Anyi Rao, and Maneesh Agrawala.
\newblock Adding {Conditional} {Control} to {Text}-to-{Image} {Diffusion} {Models}, 2023{\natexlab{b}}.
\newblock arXiv:2302.05543.

\bibitem[Zheng et~al.(2024)Zheng, Zhou, Li, Qi, Shan, and Li]{zheng_layoutdiffusion_2024}
Guangcong Zheng, Xianpan Zhou, Xuewei Li, Zhongang Qi, Ying Shan, and Xi Li.
\newblock {LayoutDiffusion}: {Controllable} {Diffusion} {Model} for {Layout}-to-image {Generation}, 2024.
\newblock arXiv:2303.17189 [cs].

\end{thebibliography}
}

\end{document}